\documentclass{article}
\usepackage{spconf,amsmath,graphicx}
\usepackage{cite}
\usepackage{algorithmic}
\usepackage{textcomp}
\usepackage{amsthm}
\usepackage{xcolor}
\usepackage{caption}
\usepackage{subfigure}
\usepackage{algorithm}
\usepackage{booktabs} 
\usepackage{multirow}
\usepackage{threeparttable}
\usepackage{placeins}
\usepackage{stfloats}
\usepackage{float}
\author{Anonymous}

\title{Robust Neural Architecture Search}
\name{Xunyu Zhu$^{\star \dagger}$ \qquad Jian Li$^{\star}$ \qquad Yong Liu$^{\ddagger}$ \qquad Weiping Wang$^{\star \dagger}$ }

\address{$^{\star}$	Institute of Information Engineering, Chinese Academy of Sciences, Beijing, China \\
	$^{\dagger}$ School of Cyber Security, University of Chinese Academy of Sciences, Beijing, China \\
	$^{\ddagger}$ Gaoling School of Artificial Intelligence, Renmin University of China, Beijing, China}

\begin{document}
%
\maketitle
\begin{abstract}
Neural Architectures Search (NAS) becomes more and more popular over these years. However,  NAS-generated models tends to suffer greater vulnerability to various malicious attacks. Lots of robust NAS methods leverage adversarial training to enhance the robustness of NAS-generated models, however, they neglected the nature accuracy of NAS-generated models.  In our paper, we propose a novel NAS method, Robust Neural Architecture Search (RNAS). To design a regularization term to balance accuracy and robustness, RNAS generates architectures with both high accuracy and good robustness. To reduce search cost, we further propose to use noise examples instead adversarial examples as input to search architectures.  Extensive experiments show that RNAS  achieves state-of-the-art (SOTA) performance on both image classification and adversarial attacks, which illustrates the proposed RNAS achieves a good tradeoff between robustness and accuracy.
\end{abstract}
\section{Introduction}


In recent years, neural architecture search (NAS) is proposed to design architectures automatically.  NAS consists of three parts: search space, search strategy, and evaluation strategy. The search space is a  candidate neural network set. The search strategy defines the way to explore search space. The evaluation strategy evaluates  the performance of the subnet. The NAS samples candidate  architectures from the search space according to the search strategy, and evaluates the performance of the selected candidate network architectures by using the evaluation strategy. According to the results of the evaluation strategy, NAS optimizes candidate network architectures  until finding the best architecture. Recently, lots of popular NAS methods emerges, such as DARTS \cite{liu2018darts, zhu2021operation, zhu2023improving}, SPOS \cite{guo2020single}, and so on. However, the performance of architectures searched by these popular NAS methods is not often good enough with bad robustness \cite{pang2021security}. 


On the other hand, adversarial defense is proposed to enhance the robustness of the network. Recently, adversarial training becomes the mainstream method of adversarial defense. It generates adversarial examples as input to train the network to enhance  robustness of the network. Now, PGD \cite{madry2018towards} is a mainstream method to generate adversarial examples. However, Tsipras et al. \cite{tsipras2018robustness} proposes that when we try to enhance the robustness of a network, the accuracy of the network will decline, there is a tradeoff between robustness and accuracy. To improve both accuracy and robustness, it is a good idea to   design new networks with better performance.



In our paper, we try to combine NAS with adversarial training to search for architectures with high accuracy and good robustness. Different from previous works \cite{guo2020meets,li2021neural} that they only leverage adversarial examples to train NAS to generate the architectures with good robustness and neglect the accuracy of the architectures, we design a regularization term to take both accuracy and robustness into consideration. The regularization term is to compute the correlation of the output under natural examples and adversarial examples.  We call our method RNAS simply. Our method includes two sub-methods, i.e., RNAS-max and RNAS-uniform. RNAS-max first generates adversarial examples as input to train NAS, it can make NAS search for architectures with better performance. However, RNAS-max makes NAS a three-level optimization problem, the computational complexity is very high. RNAS-uniform tries to solve the problem from an optimistic view. RNAS-uniform samples random noise from the perturbed set  and uses these noise examples to train NAS. We think that the subnets with good performance will have bigger architecture weights $\alpha$ to make the supernet have a better performance on noise examples. RNAS-uniform makes NAS still a bi-level optimization problem, and it only needs to spend a little cost to generate noise examples. 

Lots of experimental results demonstrate that RNAS can search architectures with good robustness and high accuracy. On CIFAR-10, RNAS-max achieves 2.65\% accuracy, and RNAS-uniform achieves 2.60 \% accuracy. Under the FGSM attack, RNAS-max still gets 53.67\% robust accuracy and RNAS-uniform still gets 53.74\% robust accuracy. When we use adversarial training to train the architectures searched by RNAS, the robust accuracy is also superior to DARTS and some of its variants. 

%

\section{Related Work}


\subsection{Neural Architecture Search }

In the past few years, NAS has become more and more popular because it can take a little  consumption cost to design neural network architectures with good  performance\cite{simonyan2014very}. However, NAS-generated models tend to suffer greater vulnerability to various malicious attacks (e.g., adversarial evasion, model poisoning, and functionality stealing) \cite{pang2021security}. Lots of methods have been proposed to improve the robustness of NAS. RobNet \cite{guo2020meets} proposed to leverage adversarial training to improve the robustness. RACL \cite{dong2020adversarially} proposed to reduce the Lipschitz constant to improve the robustness of NAS.  These methods are either too complex or  only concerned  on robustness. In our paper, we introduce a simple regularizer to consider the correlation between the output of the supernet on natural data and adversarial data and use the regularizer to balance  accuracy and robustness. 

\subsection{Adversarial Attacks and Defenses}

Adversarial attacks are to generate adversarial data to attack models. There are some kinds of adversarial attacks, including evasion attacks \cite{madry2019deep}, poisoning attacks \cite{biggio2013poisoning}, and so on. In the paper, we mainly discuss evasion attacks. Evasion attacks attach imperceptible perturbation $\delta$ on input $x$ to generate an adversarial input $x + \delta$. Then, the adversarial input $x + \delta$ is used to deceive the model to miscategorize input $x$ into other classes. Thus, it is important to generate perturbation $\delta$, and the problem can be formulated as follows:
\begin{equation}
	\min _{\delta \in \mathcal{B}} \ell(f(x+\delta), t),
\end{equation}
where $\mathcal{B}$ is represented as the perturbation set.

The aim of adversarial defense methods is to enhance the robustness of the models so that defense adversarial attacks, and there are lots of adversarial defense methods. Adversarial training is a kind of adversarial defenses methods. Adversarial training is to generate adversarial examples and use these adversarial examples to train the model to enhance its robustness of the model. PGD \cite{madry2018towards} and its variant \cite{zhang2020attacks,shafahi2019adversarial} are used to solve the min-max problem and get  good results. However, Tsipras et al. \cite{tsipras2018robustness} proposes that there is a trade-off between accuracy and robustness, thus,  high accuracy and good robustness cannot be both achieved. To solve the problem, it is a good idea to use NAS to design the networks with high accuracy and robustness. Lots of works \cite{guo2020meets,li2021neural} have been proposed to prove the feasibility of this idea. In our paper, we also try to combine NAS and adversarial training to search architectures with good performance.

\section{Methodology}

\subsection{Preliminary}
\label{preliminary}

At first, we make a brief introduction about adversarial training and neural architecture search(NAS).

\textbf{Adversarial Training.}  Deep neural networks (DNNs) are vulnerable to perturbs on examples, and  the performance of these models becomes bad under attack. Thus, adversarial training \cite{szegedy2013intriguing} is proposed to enhance the robustness of DNNs, it uses adversarial examples instead of natural examples as input to train DNNs. The problem of  adversarial training can be formulated as a minimax optimization problem:
\begin{equation}
	\mathop{min} \limits _\theta \mathop{E}\limits _{(x,y)\sim \mathcal{D}}[\mathop{max}\limits _{\left \| x'-x \right \|\le \epsilon  }\mathcal{L}(f_\theta (x'),y) ],
\end{equation}
where $f_\theta (\cdot)$ denotes the model, $D$ denotes the data distribution and $\left \| x'-x \right \|\le \epsilon  $  defines the set of allowed perturbation inputs. Then, Madry et al. \cite{madry2018towards} leverages Projected Gradient Descent (PGD) to solve the  problem to generate adversarial examples and use these examples as input to train the model to enhance the robustness. In our work, we also leverage adversarial examples to  search those robust architectures.

\textbf{Neural Architecture Search.} Neural Structure Search (NAS) is a technology to design neural network architecture automatically. NAS consists of three parts: search space, search strategy, and performance estimation. The search space includes a set of candidate neural networks that can be searched. The search strategy defines the method to find the optimal candidate network. The performance estimation strategy aims to  evaluate the performance of the sampled network. The main purpose of NAS is to find the optimal network from the search space, and it can be represented as a bi-level optimization problem:
\begin{equation}
	\begin{aligned}
		&\mathop{min} \limits _{\alpha}   \mathcal{L}_{val}(w^*(\alpha ), \alpha)     \\
		&\text { s.t. } w^{*}(\alpha)=\mathop{argmin}\limits_{w} \mathcal{L}_{\text {train }}(w, \alpha),
		\label{NAS}
	\end{aligned}
\end{equation} 
where $\alpha$ denotes the candidate network and $w$ denotes the parameters of the networks. Recently, Differentiable Architecture Search (DARTS) \cite{liu2018darts} as a very popular NAS  method due to low computing consumption and extremely fast search speed. In our work, we will leverage DARTS as a representative NAS method to search robust architectures.

\subsection{Robust Neural Architecture Search}
\label{irNAS}

The architectures searched by NAS under nature examples have  the better accuracy, however, they are not robust and vulnerable to adversarial examples. On the other hand, the architectures searched by NAS under adversarial examples have high robustness, but there was a decrease in their accuracy. Our aim is to search architectures with high accuracy and good robustness.  

\begin{figure}[]
	\vskip 0.2in
	\begin{center}
		\centerline{\includegraphics[width=\columnwidth, height=4cm]{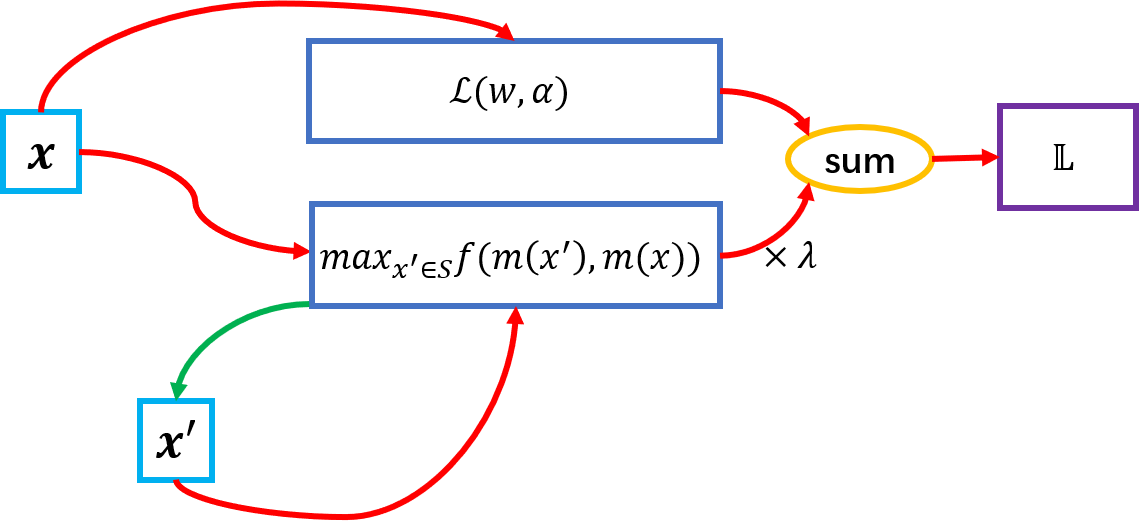}}
		\caption{Main framework of RNAS-max.}
		\label{NAS-max}
	\end{center}
	\vskip -0.2in
\end{figure}

In our paper, we  use model prediction performance  under natural examples  as the  metric to search the architectures with good accuracy. On the other hand, we propose a regularization term to search the robust architectures, i.e., maximize adversarial output similarity under nature examples and adversarial examples.   Thus, the NAS optimization problem can be formulated as follows:
\begin{equation}
	\begin{aligned}
		&\mathop{min} \limits _{\alpha}   \mathcal{L}_{val}(w^*(\alpha ), \alpha) + \lambda   \mathop{argmax} \limits_ {x' \in S}f(m(x'), m(x))   \\
		&\resizebox{.9\hsize}{!}{$\text { s.t. } w^{*}(\alpha)=\mathop{argmin}\limits_{w} [\mathcal{L}_{\text {train }}(w, \alpha) + \lambda   \mathop{argmax} \limits_ {x' \in S}f(m(x'), m(x))],$}
		\label{bdy_opt}
	\end{aligned}
\end{equation} 
where $m$ denotes the supernet and $\lambda$ as a regularization coefficient controls the tradeoff between robustness and accuracy. When $\lambda$ becomes larger, more attention will be paid to robustness than accuracy. We call our method  RNAS-max simply.  


Fig. \ref{NAS-max} introduces the main framework of RNAS-max. PGD \cite{madry2018towards}  is a mainstream optimization method to generate adversarial examples. Thus, we also leverage PGD to optimize RNAS-max, and the algorithm of RNAS-max is represented as Alg. \ref{alg:rNAS_max}. At first, we initialize architecture parameters $\alpha$ and network parameters $w$. Then, we use valid data as input and leverage PGD to optimize  $\mathop{argmax} \limits_ {x' \in S}f(m(x'), m(x))$ to generate adversarial valid examples. After adversarial valid examples are generated, we leverage adversarial valid examples and natural valid examples as input to put in the supernet to compute their outputs. Then, we compute the correlation of two outputs and nature valid loss and take the sum of them as a robust valid loss. We leverage Adam to optimize robust valid loss to get the optimal architecture parameters $\alpha$. Similar to the DARTS, we alternate optimizing the architecture parameters $\alpha$ and the network parameters $w$. Thus, we use training data as input to generate adversarial training examples  by optimizing  $\mathop{argmax} \limits_ {x' \in S}f(m(x'), m(x))$. Then, we use adversarial training examples and natural training examples as the input of the inner optimization problem (Eq. \ref{bdy_opt}) and use SGD to optimize the inner optimization problem to get the optimal network parameters $w$. We use this way to optimize the problem (Eq. \ref{bdy_opt})  several steps until get the optimal architecture parameters $\alpha$ and the network parameters $w$. Finally, we derive the optimal subnet with good robustness and high accuracy.

\begin{algorithm}[h]
	\caption{RNAS-max}
	\label{alg:rNAS_max}
	\begin{algorithmic}
		\STATE {\bfseries Input:} supernet $m$, similarity metric $f$
		\STATE Initialize architecture parameters $\alpha$ and network parameters $w$.
		\WHILE {not converged}
		\STATE Generate adversarial valid examples by using PGD to optimize $\mathop{max} \limits_ {x' \in S}f(m(x'), m(x))$ where $x$ is valid data;
		\STATE Update $\alpha$ by optimizing $\mathop{min} \limits _{\alpha}   \mathcal{L}_{val}(w^*(\alpha ), \alpha) + \lambda  \mathop{argmax} \limits_ {x' \in S}f(m(x'), m(x))$;
		\STATE Generate adversarial training examples by using PGD to optimize $\mathop{max} \limits_ {x' \in S}f(m(x'), m(x))$ where $x$ is training data;
		\STATE Update $w$ by optimizing $\mathop{argmin}\limits_{w} [\mathcal{L}_{\text {train }}(w, \alpha) + \lambda  \mathop{argmax} \limits_ {x' \in S}f(m(x'), m(x))]$;
		\ENDWHILE
	\end{algorithmic}
\end{algorithm}

\subsection{Noise Examples on RNAS}

Section \ref{irNAS} proposes to add a regularization term to the bi-level optimization problem of NAS to generate neural network architectures with higher quality. Although the method is effective, it is not efficient enough. Eq. \ref{bdy_opt} becomes a three-level optimization problem, and the computational complexity of the three-level optimization problem (Eq. \ref{bdy_opt}) is far higher than the bi-level optimization problem (Eq. \ref{NAS}). It will take several times longer than standard DARTS to generate the optimal subnet. The problem is a serious obstacle to the practical application of our method. In our paper, we try to generate some noise examples instead of adversarial examples to train RNAS for reducing consumption cost. 

\begin{figure}[]
	\vskip 0.2in
	\begin{center}
		\centerline{\includegraphics[width=\columnwidth, height= 3cm]{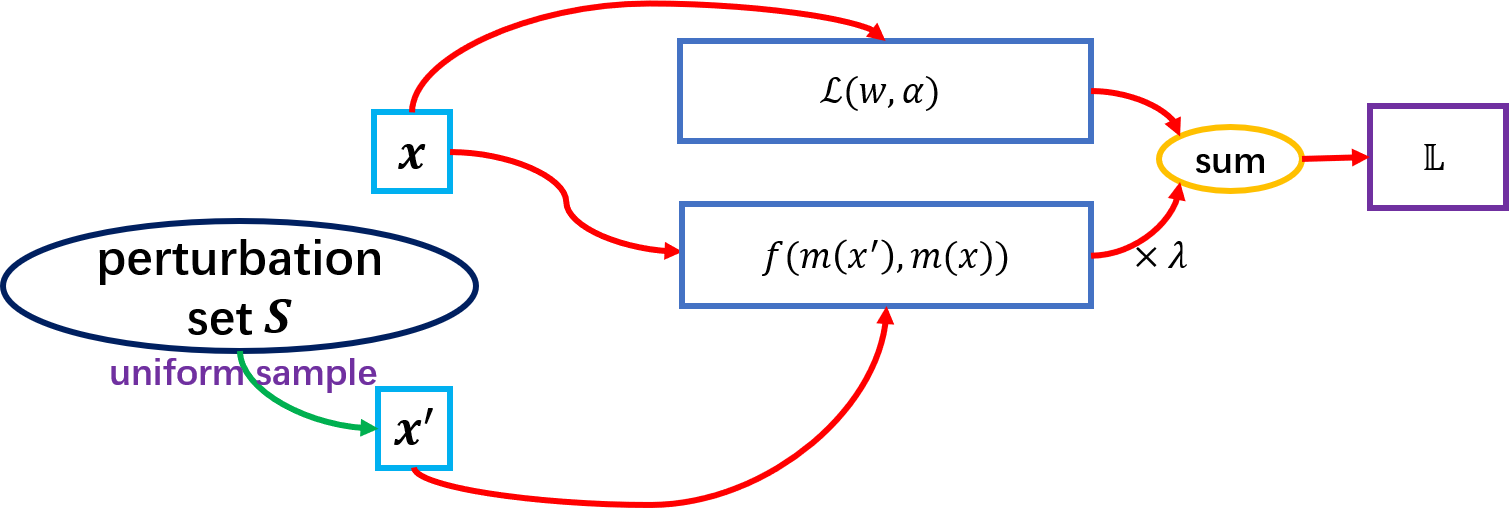}}
		\caption{Main framework of RNAS-uniform.}
		\label{NAS-uniform}
	\end{center}
	\vskip -0.2in
\end{figure}

We think that robust models are also insensitive to random noise in samples, and the supernet is the set of lots of subnets. To make the supernet insensitive to random noise in samples, these robust subnets will have bigger architecture weights than others.  We sample some random examples from the perturbation set as noise examples instead of  adversarial examples.  We call the method RNAS-uniform simply. Then, the RNAS-uniform can be formulated as follows:
\begin{equation}	
	\begin{aligned}	
		&\mathop{min} \limits _{\alpha}   \mathcal{L}_{val}(w^*(\alpha ), \alpha) + \lambda   \mathop{U} \limits_ {x' \in S}f(m(x'), m(x))  \\
		&\resizebox{.9\hsize}{!}{$\text { s.t. } w^{*}(\alpha)=\mathop{argmin}\limits_{w} [\mathcal{L}_{\text {train }}(w, \alpha) + \lambda  \mathop{U} \limits_ {x' \in S}f(m(x'), m(x))],$}
		\label{rNAS-uniform}
	\end{aligned}
\end{equation} 
where $x'$ is represented as noise examples, and $U$ is represented to sample a perturbation example  uniformly from the perturbation set. Eq. \ref{rNAS-uniform} becomes an bi-level optimization problem, and we only need to spend a little extra consumption cost to generate noise examples. Although we leverage noise examples to train NAS to generate architectures with worse robustness and lower accuracy than adversarial examples, it can reduce search time a lot. 

\begin{algorithm}[tb]
	\caption{RNAS-uniform}
	\label{alg:rNAS_uniform}
	\begin{algorithmic}
		\STATE {\bfseries Input:} supernet $m$, similarity metric $f$
		\STATE Initialize architecture parameters $\alpha$ and network parameters $w$.
		\WHILE {not converged}
		\STATE Generate perturbation valid examples by sampling at random from the perturbation valid set;
		\STATE Update $\alpha$ by optimizing $\mathop{min} \limits _{\alpha}   \mathcal{L}_{val}(w^*(\alpha ), \alpha) + \lambda  \mathop{U} \limits_ {x' \in S}f(m(x'), m(x))$;
		\STATE Generate perturbation training examples by sampling at random from the perturbation training set;
		\STATE Update $w$ by optimizing $\mathop{argmin}\limits_{w} [\mathcal{L}_{\text {train }}(w, \alpha) + \lambda   \mathop{U} \limits_ {x' \in S}f(m(x'), m(x))]$;
		\ENDWHILE
	\end{algorithmic}
\end{algorithm}

Fig. \ref{NAS-uniform} introduces the main framework of RNAS-uniform, and  the algorithm of RNAS-max is represented as Alg. \ref{alg:rNAS_uniform}. At first, we generate perturbation valid examples by sampling uniformly from the perturbation valid set. Then, we use the perturbation valid examples and the natural valid examples as input, and update architecture parameters $\alpha$ by optimizing $\mathop{min} \limits _{\alpha}   \mathcal{L}_{val}(w^*(\alpha ), \alpha) + \lambda  \mathop{U} \limits_ {x' \in S}f(m(x'), m(x))$. We also alternate  optimizing architecture parameters $\alpha$ and network parameters $w$ as   DARTS. Thus, we generate perturbation training examples by sampling uniformly from the perturbation training set. Afterwards, we use the perturbation training examples and the natural valid examples as input, and update network parameters $w$ by optimizing $\mathop{argmin}\limits_{w} [\mathcal{L}_{\text {train }}(w, \alpha) + \lambda   \mathop{U} \limits_ {x' \in S}f(m(x'), m(x))]$. We use the way to optimize the problem (Eq. \ref{rNAS-uniform}) several steps. Finally, we derive the optimal subnet based on architecture parameters. 

\section{Experiments}



%

\subsection{Architecture Search}

CIFAR-10 includes 50K training images and 10K testing images, and each image's spatial resolution is $32 \times 32$. These images belong to 10 classes. When searching the architecture, we divide these training images into two subnets, i.e., the training subnet and the valid subnet. The training subnet is used to optimize network parameters, and the valid subnet is leveraged to optimize architecture parameters. 

DARTS is one of the most popular NAS methods, and  we use DARTS as a representative method to combine our method to search the architecture with good performance. Similar to standard DARTS, we aim to search for a normal cell and a reduction cell to build a network with good performance on the image classification task.  Our method has two submethods, i.e., RNAS-max and RNAS-uniform, and we  introduce the experiment settings of the two submethods as follows.

	%
	%
	%
%
	%
	%
	%

We introduce the experiment setting of RNAS-max at first. Most of our experiment settings are consistent with DARTS, and we make a few modifications to others. We run rNAS for a total of 50 epochs to search the architecture, namely total epochs $T$ for search is 50. The batch size $m$ of our experiment is 64. The regularization coefficient $\lambda  $ is set as 1.0. The neural network at initialization does not contain valuable information, so we first warm up the neural network for 15 epochs, i.e., epochs $W$ for warm-up is 15. We use PGD to generate adversarial examples. In our experiment settings, we use 10-PGD to generate adversarial examples, i.e., num steps $K$ is 10. Further, the step size $\eta$ of 10-PGD is 0.003, and the total perturbation scale $\epsilon$ is 0.031. We use momentum SGD to update network parameters $w$, and its initial learning rate is 0.025, momentum is 0.9, and weight decay factor 3e-4. Then, we use Adam to update architecture parameters $\alpha$, and its initial learning rate is 3e-4, momentum is $(0.5, 0.999)$, and weight decay factor is 1e-3. We spend 4.3 GPU days generating the architecture.

\begin{table*}[ht]
	\centering
	\small
	\caption{Comparison with state-of-the-art image classifiers on CIFAR-10.}
	\label{acc_c10}
	
	\begin{tabular}{@{}ccccc@{}}
		\toprule
		\textbf{Architecture} &
		\begin{tabular}[c]{@{}c@{}} \textbf{Test Err.}\\ (\textbf{\%})\end{tabular} &
		\begin{tabular}[c]{@{}c@{}}\textbf{Params}\\ \textbf{(M)}\end{tabular} &
		\multicolumn{1}{l}{\begin{tabular}[c]{@{}l@{}}\textbf{Search Cost}\\ \textbf{(GPU-days)}\end{tabular}} &
		\begin{tabular}[c]{@{}c@{}}\textbf{Search} \\ \textbf{Method}\end{tabular} \\ \midrule
		DenseNet-BC \cite{huang2017densely}    & 3.46  & 25.6 & -    & manual    \\ \midrule
		NASNet-A \cite{zoph2018learning} & 2.65     & 3.3  & 1800 & RL        \\
		AmoebaNet-A \cite{real2019regularized}    & 3.34$\pm$0.06 & 3.2  & 3150 & evolution \\
		AmoebaNet-B \cite{real2019regularized}    & 2.55$\pm$0.05 & 2.8  & 3150 & evolution \\
		PNAS  \cite{liu2018progressive}        & 3.41$\pm$0.09 & 3.2  & 225  & SMBO      \\
		ENAS \cite{pham2018efficient}          & 2.89          & 4.6  & 0.5  & RL        \\ \midrule
		DARTS ($1^{\text{st}}$ order)  \cite{liu2018darts} & 3.00$\pm$0.14 & 3.3  & 0.4  & gradient  \\
		DARTS ($2^{\text{nd}}$ order) \cite{liu2018darts} & 2.76$\pm$0.09 & 3.3  & 1    & gradient  \\
		SNAS (mild)  \cite{xie2018snas}      & 2.98          & 2.9  & 1.5  & gradient  \\
		ProxylessNAS  \cite{cai2018proxylessnas}     & 2.08          & -    & 4    & gradient  \\
		P-DARTS  \cite{chen2019progressive}       & 2.5           & 3.4  & 0.3  & gradient  \\
		PC-DARTS  \cite{xu2019pc}     & 2.57$\pm$0.07 & 3.6  & 0.1  & gradient  \\
		SDARTS-RS \cite{2020arXiv200205283C}     & 2.67$\pm$0.03 & 3.4  & 0.4  & gradient  \\
		GDAS  \cite{dong2019searching}     & 2.93          & 3.4  & 0.3  & gradient  \\
		R-DARTS (L2) \cite{zela2019understanding}       & 2.95$\pm$0.21 & -    & 1.6  & gradient  \\
		SGAS (Cri 1. avg)  \cite{li2020sgas}    & 2.66$\pm$0.24 & 3.7  & 0.25 & gradient  \\
		DARTS-PT \cite{wang2021rethinking}        & 2.61$\pm$0.08 & 3.0  & 0.8  & gradient  \\ \midrule
		RNAS-max     & 2.65  & 3.4  & 4.3   & gradient  \\ 
		RNAS-uniform     & 2.60  & 3.4 & 0.5   & gradient  \\ \bottomrule 	
	\end{tabular}
\end{table*}

The experiment settings of RNAS-uniform are also consistent with DARTS and RNAS-max mostly. The total epochs $T$ of RNAS-uniform are also 50, including 15 epochs $W$ for warm\_up. The batch size $m$ and regularization coefficient $\lambda$ of RNAS-uniform are also consistent with RNAS-max, i.e., 64 and 1.0. We sample uniformly  noise examples from the perturb set, and the perturbation bound $\epsilon$ is 0.031.  The optimizers to optimize architecture parameters $\alpha$ and network parameters $w$ are consistent with RNAS-max, and the hyperparameters of the two optimizers are also the same as RNAS-max. We spend 0.5 GPU days to generate our architecture, and the search time is only slightly slower than DARTS.


\subsection{Architecture Evaluation on CIFAR-10}

We  use these cells searched by RNAS to build  networks and evaluate the performance of these networks on CIFAR-10.   We use 20 cells to build a network, including two reduction cells located at 1/3 and 2/3 of the total depth of the network.  We run the training process 600 epochs, and the batch size of our training set is 96. We also use SGD to optimize the network, and the initial learning rate is 0.025. We also cutout and auxiliary towers to help us train the networks; the length of the cutout is 16, the weight of auxiliary towers is 0.4, and the probability of path dropout is 0.3.

Table \ref{acc_c10} summarizes evaluation results and comparison with state-of-the-art approaches. Table  \ref{acc_c10} shows that RNAS-max achieves a 2.65\% test error on CIFAR10 with a search cost of 4.3 GPU-days. Although the search time of RNAS-max is beyond DARTS a lot, the performance is far better than standard DARTS; the params of the architecture generated by RNAS-max is 3.4M, and it is a little larger than the architecture searched by DARTS and smaller than the architecture searched by some variants of DARTS, e.g., SGAS \cite{li2020sgas}, PC-DARTS  \cite{xu2019pc}. RNAS-uniform achieves a 2.60\% test error on CIFAR-10 with a search cost of 0.5 GPU-days. The performance of RNAS-uniform is far better than standard DARTS, while the search time of RNAS-uniform is 0.5 GPU days, less than DARTS ($2^{\text{nd}}$ order), and a little more than DARTS ($1^{\text{st}}$ order).  The parameters of the architecture searched by RNAS-uniform are also 3.4M, and the architecture is lightweight enough to be run on mobile devices. In general, RNAS makes a big improvement on DARTS, and can search good architectures with high accuracy and good robustness.

\begin{table}[ht]
	\centering
	\scriptsize
	\caption{Evaluation of robust accuracy on CIFAR-10 under adversarial attacks.}
	\setlength{\tabcolsep}{5pt}
	\vspace{5pt}
	\renewcommand{\arraystretch}{1.13}
	\begin{tabular}{l|r|ccc|c}
		\toprule
		& & \multicolumn{3}{c|}{\textbf{Adversarially Trained}} & \multicolumn{1}{c}{\textbf{Standard Trained}} \\ 
		\textbf{Model} & \textbf{Params} & \textbf{Clean} & \textbf{FGSM} & \textbf{PGD$^{20}$ } &  \textbf{FGSM} \\
		\bottomrule \toprule
		ResNet-18 \cite{he2016deep} & 11.2M & 84.09\% & 54.64\% & 45.86\%  & 50.71\% \\
		DenseNet-121 \cite{huang2017densely}  & 7.0M & 85.95\% & 58.46\% & 50.49\%   & 45.51\% \\
		\midrule
		DARTS \cite{liu2018darts} & 3.3M & 85.17\% & 58.74\% & 50.45\%  & 50.56\% \\
		PDARTS \cite{chen2019progressive} & 3.4M & 85.37\% & 59.12\% & 51.32\%   & 54.51\% \\
		RobNet-free \cite{guo2020meets} & 5.6M & 85.00\% & 59.22\% & 52.09\%    & 36.99\% \\
		RACL \cite{dong2020adversarially} & 3.6M & 84.63\% & 58.57\% & 50.62\%  & 52.38\% \\
		\midrule
		RNAS-max & 3.4M & \textbf{86.30\%} & \textbf{59.59\%} & \textbf{52.65\%}    & \textbf{53.67\%} \\
		RNAS-uniform & 3.4M & \textbf{85.42\%} & \textbf{55.36\%} & \textbf{47.52\%}    & \textbf{53.74\%} \\
		\bottomrule
	\end{tabular}
	\vspace{-5pt}
	\label{table:whitebox}
\end{table}

\subsection{Robustness of Architecture}

We evaluate the robustness of architectures that are standard and adversarially trained on CIFAR-10 by using lots of adversarial attacks.  Standard training is to leverage natural examples to train the networks. The experiment settings of standard training are the same as DARTS. Adversarial training is to use adversarial examples to train the networks. In our paper, we use 7-step PGD to train the networks; the step size is 0.01 and the perturbation scale is 0.031.  We use FGSM to attack standard-trained architectures; FGSM and PGD$^{20}$ are used to attack adversarially-trained architectures. Table \ref{table:whitebox} shows the evaluation results on CIFAR-10 under adversarial attacks. RNAS-max achieves good robust accuracy by standard training or adversarially training, it gets the highest robust accuracy than other methods. However, RNAS-uniform has bad performance on robustness, and it means that  RNAS-uniform  sacrifices the robustness of the architecture and has a little effect on accuracy by using noise examples as input.  In the further, we seek a better method to improve the robustness and accuracy of the architectures searched by NAS with low consumption cost.

\section{Conclusion}

In our works, we propose a novel NAS method, RNAS. RNAS designs a regularization term to balance accuracy and robustness during the search process of NAS. To reduce the search time of RNAS, we try to use noise examples instead of adversarial examples to train NAS. Experiment results shows the RNAS is effective and achieve SOTA performance on image classification and adversarial attacks. In future, we will try to design a more efficient NAS method to search architectures with better performance.

\bibliographystyle{IEEEbib}
\bibliography{refs}

\end{document}